\documentclass[runningheads,a4paper]{llncs}

\usepackage{amssymb}
\usepackage{graphicx}
\usepackage{array}
\usepackage{hyperref}
\usepackage{multirow}
\usepackage{rotating}

\usepackage{url}
\newcommand{\keywords}[1]{\par\addvspace\baselineskip
\noindent\keywordname\enspace\ignorespaces#1}

\begin{document}

\mainmatter 
\title{Visual saliency estimation by integrating features using multiple kernel learning}
\titlerunning{Visual saliency estimation by integrating features using MKL}

%
%
\author{Yasin Kavak \and Erkut Erdem \and Aykut Erdem}
\institute{Hacettepe University, Department of Computer Engineering\\Ankara, TR-06800, Turkey\\
{\tt\small \{n10164820,erkut,aykut\}@cs.hacettepe.edu.tr}}
%


%
%

\toctitle{Lecture Notes in Computer Science}
\tocauthor{Authors' Instructions}
\maketitle

\begin{abstract}
In the last few decades, significant achievements have been attained in predicting where humans look at images through different computational models. However, how to determine contributions of different visual features to overall saliency still remains an open problem. To overcome this issue, a recent class of models formulates saliency estimation as a supervised learning problem and accordingly apply machine learning techniques. In this paper, we also address this challenging problem and propose to use multiple kernel learning (MKL) to combine information coming from different feature dimensions and to perform integration at an intermediate level. Besides, we suggest to use responses of a recently proposed filterbank of object detectors, known as ObjectBank, as additional semantic high-level features. Here we show that our MKL-based framework together with the proposed object-specific features provide state-of-the-art performance as compared to SVM or AdaBoost-based saliency models. 
\end{abstract}
\keywords{visual saliency, multiple kernel learning, feature integration, object-based attention}

\section{Introduction}
Enabling computers to understand the external world around them has attracted the interest of many researchers in different disciplines. A primary source of inspiration for solving this hard problem is our visual system, and some researcher have come up with elegant solutions which imitate certain mechanisms present in the human visual system. In that respect, computational models of visual attention have gained popularity for the last decades. As being one of the most important characteristics of human vision, attentional mechanisms manage limited processing resources by selecting the most interesting or salient parts of a scene while ignoring other stimuli. Accordingly, they help carry out certain visual tasks such as object recognition and scene understanding with ease.

Motivated by the theories of preattentive vision, namely the Feature Integration Theory~\cite{Treisman:CogPsych1980} or the Guided Search Model~\cite{Wolfe:PsychBul1994}, a large number of computational methods have been proposed in the past decade with the aim of predicting where people are likely to look in images. These studies introduce different saliency models which are grounded in different computational frameworks. Besides helping understanding the human visual system, they also have many applications such as scene classification~\cite{Siagian:TPAMI2007}, object recognition~\cite{Gao:TPAMI2009}, and object tracking~\cite{Butko:ICRA2008}.

In computer vision literature, the interest in visual salience methods dates back to the late 80's~\cite{Koch:HumNeurobiol:1985,Clark:ICCV1988,Chapman:MIT1990,Tsotsos:AI1995,Jagersand:ICCV1995,Niebur:NIPS1996}. As a basic implementation of the approach of Koch and Ullman~\cite{Koch:HumNeurobiol:1985}, the Itti-Koch model~\cite{Itti:TPAMI1998} is considered one of the earliest modern saliency models. It attempts to saliency estimation by computing center-surround differences for a set of features by means of Difference of Gaussians and then linearly combines them into a final saliency map, mimicking the workings of retinal receptive fields. Another popular model is Harel et al.'s the graph-based visual saliency (GBVS) model~\cite{Harel:NIPS2006}. It also computes some feature maps at multiple scales but represents them as fully connected graphs, and then estimates the resulting saliency map from these graphs using a graph-theoretic approach. Other alternatives include frequency domain-based approaches~\cite{Hou:CVPR2007,Achanta:ICIP2009} and probabilistic formulations~\cite{Torralba:JOptSocAmA2003,Bruce:NIPS2005,Torralba:PsychRev2006,Zhang:JoV2008}. Please refer to~\cite{Borji:TPAMI2013} for a more detailed review.

Most of the existing models consider the bottom-up strategy suggested in the Itti's model~\cite{Itti:TPAMI1998} in which center-surround differences of various features at multiple scales are computed for each feature dimension and a final saliency map is obtained by combining these individual feature maps in a linear manner after a normalization step. This approach suggests that every visual feature is equally important for saliency estimation so that the integration step can be carried out in a straightforward way. However,  how different feature dimensions contribute to the overall saliency is still an open question. In recent years, a number of computational models~\cite{Liu:CVPR2007,Judd:ICCV2009,Zhao:JoV2011,Zhao:JoV2012,Borji:CVPR2012} try to overcome the \textit{feature integration} problem by posing saliency estimation as a supervised learning problem where optimal values are found for different feature maps or a combination of them. 

\begin{figure}[!h]
\centering
\includegraphics[height=0.195\linewidth]{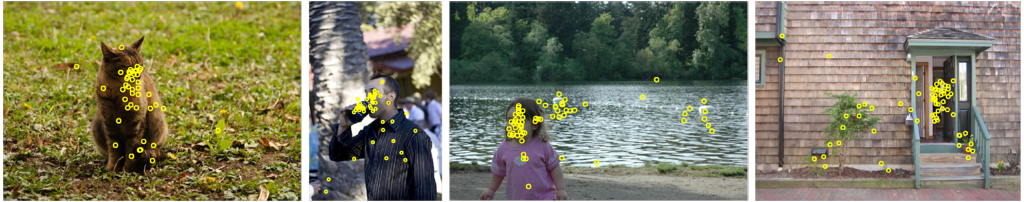} 
\caption{Sample images in which viewers mostly attended to certain objects such as a cat, a bottle, human faces and a person (taken from MIT1003 dataset~\cite{Judd:ICCV2009}).}
\label{fig:highlevel}
\end{figure}

In this study, we address the feature integration problem in visual saliency estimation. The proposed model employs a supervised approach to classify image pixels as salient or non-salient. Our main contributions are two-folds. First, we propose to use \textit{multiple kernel learning (MKL)} paradigm~\cite{Gonen:JMLR2011} to learn a classifier which combines information coming from multiple sources at an intermediate step and in effect integrates features in a more effective way. More specifically, we employ certain variants of MKL algorithms which provide combination schemes that are more complex than ordinary linear summation and achieve better prediction performances compared to previous learning-based formulations. Second, motivated by an increasing number of studies which suggest that humans generally attend to objects in a scene~\cite{Scholl2001,ElazaryJoV2008,EinhauserJoV2008,NuthmannJoV2010} (See Fig.~\ref{fig:highlevel}), we suggest to use a recently suggested large filter bank of object detectors called ObjectBank~\cite{Li-ObjectBank:NIPS2010} as high-level features for saliency prediction. Here, it is important to note that object-level feature maps have been previously used by some previous models~\cite{Cerf:JoV2009,Judd:ICCV2009,Goferman:CVPR2010,Zhao:JoV2012,Borji:CVPR2012} but they are limited to only a very small set of hand-picked object classes like faces, pedestrians, or cars. 


\section{Learning visual saliency}
\label{sec:learning_saliency}
In recent years, many different computational approaches to visual saliency estimation have been proposed. Starting from the seminal work by Itti, Koch and Niebur~\cite{Itti:TPAMI1998}, most of these existing methods consider a purely bottom-up strategy which traditionally follows a three-step procedure:  (a) Extract some basic low-level visual features, (b) investigate several features channels in parallel by considering center-surround differences of features to extract a separate feature map for each dimension, and (c) integrate these feature maps to come up with a master saliency map. This last feature integration step is typically carried out by weighted averaging which corresponds to linear summation. 

Within this view, the visual features used in a model play a key role as they strongly affect its prediction performance. The standard features include low-level (bottom-up) cues such as intensity, color and orientation which are used to determine regions that show different characteristics from its neighbors. Moreover, it has been shown that faces, humans, animals, text, etc. immediately grab one's attention due to top-down influences. Thus, to increase the prediction accuracy, a number of studies additionally takes into account responses of some object detectors including face or pedestrian detectors as individual feature maps~\cite{Cerf:JoV2009,Judd:ICCV2009,Goferman:CVPR2010,Zhao:JoV2012,Borji:CVPR2012}.

The aforementioned procedure has several limitations since it has many design choices and parameters. It simply falls short in explaining which features are relevant to visual saliency, how to select related parameters and how to perform the final feature integration step in a more principled way. In the past few years, a different view to visual saliency has emerged which departed from the traditional view by formulating visual saliency estimation as a supervised learning problem. In mathematical terms, this can be stated as learning a classifier $f:\mathbb{R}^D\rightarrow \{+1, -1\}$ with $D$ denoting the dimension of the features, which returns the label $+1$ for salient (fixated) points, and $-1$ for the non-salient points. The function $f$ is trained over a training set of features extracted from a group of fixated and non-fixated points. Once the training is completed, visual saliency estimation can be cast as computing the value of $f$ at each pixel of a given test image. This learning-based strategy is independent of the features used so one can make use of both low-level and high-level visual features in a unified manner. Moreover, it enables to automatically choose features relevant to visual saliency by learning specific feature weights in the integration step. 

One of the early methods, which uses this learning-based approach, was introduced by Peters and Itti~\cite{Peters:CVPR2007}. They proposed to utilize a task-dependent saliency map by training a linear least square regression classifier which takes input the gist of the scene, computed with respect to image pyramids and Fourier based features. The top-down modulation of visual attention is carried out by simply multiplying the task-based map with the bottom-up saliency map of the image. In~\cite{Zhao:JoV2011}, Zhao and Koch utilized linear regression to quantify the relevance of color, intensity, gradient and face features by means of the learned weights.

In~\cite{Kienzle:JoV2009},  to classify a pixel as salient or not, Kienzie et al. trained a support vector machine (SVM) using the local intensity values within the $13\times 13$ image patch centered on the pixel. Their choice of employing Gaussian kernels provides a non-linear mapping of input vectors into a high-dimensional feature space. Similar to the framework in~\cite{Kienzle:JoV2009}, Judd et al.~\cite{Judd:ICCV2009}  proposed another learning based model which is based on training SVMs to determine the saliency of a pixel. The authors employed both simple low-level features like intensity, orientation and color, and mid- and high-level features such as responses of horizon line detector, face and pedestrian detectors. Moreover, using a linear kernel makes the training and test phases much faster and the resulting optimal weights for the classifier directly indicates the contributions of different features to saliency estimation.

The saliency models proposed by Zhao and Koch~\cite{Zhao:JoV2012} and Borji~\cite{Borji:CVPR2012} employ the AdaBoost classification algorithm~\cite{Freund:1997} as an alternative learning paradigm to predict where humans look at images. Both models make use of a set weak classifiers to learn a final strong classifier with a better classification capability as the weighted combination of the output of these weak classifiers. An important property of these AdaBoost-based models is that they provide a nonlinear integration of different visual features.

It is also interesting to mention that there are another group of saliency studies which find the optimal weights for saliency channels specifically to find objects such Bayesian approach by Navalpakkam and Itti~\cite{Navalpakkam:CVPR2006}, heuristics-based model by Frintop~\cite{Frintrop:Bonn2005}, and GA-based technique by Borji et al.~\cite{Borji:MVA2011}. 

\section{Feature integration using multiple kernel learning}
\label{sec:mkl}
In machine learning, the support vector machine (SVM) is a class of supervised learning models which are based on the theory of structural risk minimization~\cite{Vapnik1998}. Let $\{(\mathbf{x}_i,y_i)\}^
N_{i=1}$ denote a set of $N$ independent and identically distributed training samples with $\mathbf{x}_i$ representing the $D$-dimensional input and $y_i \in \{-1,+1\}$ being its class label. Then, SVM learning finds the linear discriminant function $f(\mathbf{x})$ which has the maximum margin in the feature space given by the mapping function $\Phi:\mathbb{R}^D \rightarrow \mathbb{R}^S$. Formally, $f(\mathbf{x})$ can be trained by solving 
\begin{eqnarray} 
\textnormal{max} & & \sum_{i=1}^N \mathbf{\alpha}_i -\frac{1}{2}\sum_{i=1}^N\sum_{j=1}^N\mathbf{\alpha}_i\mathbf{\alpha}_jy_iy_j \underbrace{\langle \Phi(\mathbf{x_i}),\Phi(\mathbf{x_j})\rangle}_{k(\mathbf{x}_i,\mathbf{x}_j)} \nonumber \\
\textnormal{s.t.} & & \sum_{i=1}^N\mathbf{\alpha}_iy_i=0, \nonumber \\
& & C \geq \mathbf{\alpha}_i \geq 0 \quad \forall i
\end{eqnarray}
with $k: \mathbb{R}^D \times \mathbb{R}^D \rightarrow \mathbb{R}$ denoting the so-called the kernel function, $C$ being a trade-off parameter and $\alpha\in \mathbb{R}^N_+$ representing the vector of dual variables. With this formulation, the discriminant function can be written as: 
\begin{equation}
f(\mathbf{x})=\sum_{i=1}^N \mathbf{\alpha}_iy_ik(\mathbf{x}_i,\mathbf{x})+b.
\end{equation}

Among others, the linear kernel $\left( k(\mathbf{x}_i,\mathbf{x}_j)=\langle \mathbf{x}_i\mathbf{x}_j \rangle \right)$, the polynomial kernel $\left( k(\mathbf{x}_i,\mathbf{x}_j)=( \langle \mathbf{x}_i\mathbf{x}_j \rangle + 1) ^q \right)$, and the Gaussian Radial basis kernel $(k(\mathbf{x}_i,\mathbf{x}_j)= \textnormal{exp} (-\gamma\|\mathbf{x}_i-\mathbf{x}_j\|^2))$ functions are the most commonly used kernel functions in the literature. In practice, selecting the best kernel function $k(\cdot,\cdot)$ and finetuning its parameter need to done for every learning problem. This procedure may be cumbersome and is typically carried out by performing a cross-validation over a set of kernel choices and their parameters. 

Multiple Kernel Learning (MKL) is a recently-proposed supervised learning paradigm~\cite{Gonen:JMLR2011} which aims at eliminating the aforementioned problems of SVMs. In particular, MKL uses multiple kernels $\left\{k_m(\mathbf{x}_i^m,\mathbf{x}_j^m): \mathbb{R}^{D_m} \times \mathbb{R}^{D_m} \rightarrow \mathbb{R}\right\}_{m=1}^P$ instead of the single kernel in the SVM-based formulation to learn a more robust discriminant function where $P$ is the number of feature representations, and $\mathbf{x}_i^m \in \mathbb{R}^{D_m}$ is the representation of the $i$th instance in the $m$th feature space with $D_m$ being the dimensionality of the corresponding feature space.

The general form of MKL can be described by the following function: 
\begin{equation}
k_\eta(\mathbf{x}_i,\mathbf{x}_j)=f_\eta\left( \left\{ k_m(\mathbf{x}_i^m,\mathbf{x}_j^m)\right\}_{m=1}^P  \right) 
\end{equation}
which is used to combine the multiple kernels in a linear or nonlinear manner. 

Several different MKL formulations have proposed in the literature~\cite{Gonen:JMLR2011}, each focusing on combining different notions of similarity or different information coming from multiple sources. In our experiments, we consider three different MKL models, namely RBMKL~\cite{Cristianini2000}, NLMKL~\cite{Cortes:NIPS2010} and LMKL~\cite{Gonen:ICML2008}. RBMKL denotes the fixed-rule based MKL (RBMKL) which obtains a combined kernel by taking the multiplication of valid kernels, providing a nonlinear integration of different feature representations. NLMKL is the non-linear MKL model that learns a kernel function in the kernel space considering a polynomial combination of kernels. LMKL is the localized MKL model which learns weights in the kernel combination in a data-dependent way considering a linear integration. For detailed descriptions of these models, please refer to~\cite{Gonen:JMLR2011}.

\section{Proposed Approach}
\label{sec:approach}
\begin{figure}[!t]
\begin{center}
   \hspace*{0in}
   \vspace*{0in}
   \includegraphics[width=\linewidth]{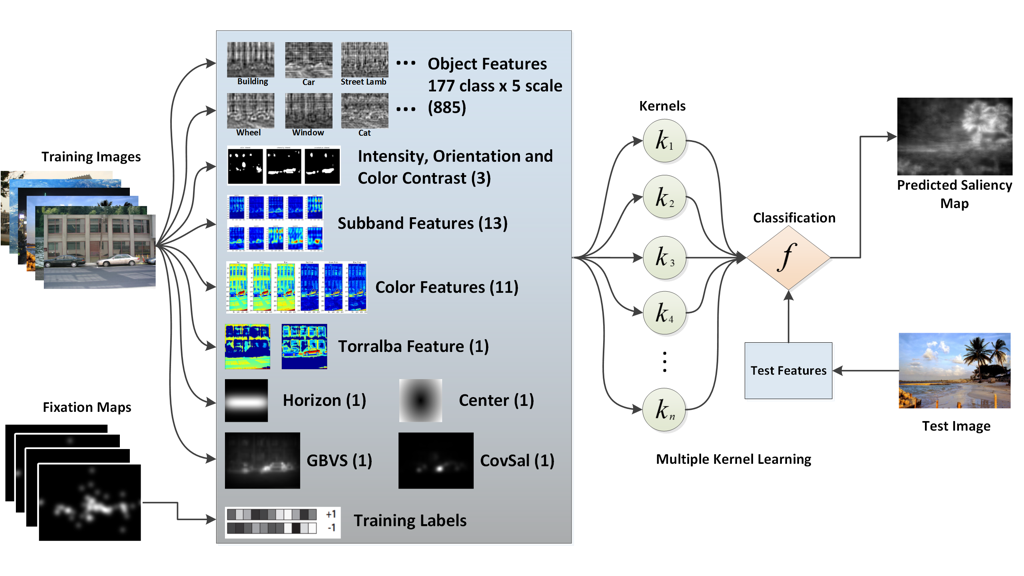}
\end{center}
   \caption{The proposed MKL-based framework for visual saliency estimation (See text for a detailed description).}
   \label{fig-mainFlow}
\end{figure}

In this paper, we propose a novel learning-based saliency model which is based on MKL framework.The classifier function is built upon using a combination of several different kernels, each highlighting a certain aspect of the pixel in a different feature space. As illustrated in Fig. \ref{fig-mainFlow}, the proposed method has \textit{training} and \textit{testing} phases. The training phase involves learning a classifier from a group of images with human eye fixation data. We first extract several low, mid and high-level features for each training image, and represent each pixel in a higher-dimensional space based on these features. Positive and negative instances correspond to fixated and non-fixated points, respectively. Once a saliency model (classifier) is trained, the testing phase includes estimating the saliency map of a given test image, which is simply carried out by first extracting the related features from each image pixel and then feeding the corresponding feature vector to our classifier. 

Our main contribution in this study is to use MKL as the underlying machine learning framework. As we mentioned in the previous section, MKL carries out feature integration in a well founded way by using different kernels and combining them so that different notions of similarity or multi-modal information are considered simultaneously. The proposed MKL based approach lies in the middle of two ends of a spectrum of approaches. One end corresponds to the case where features from different sources are concatenated and fed to a single learner, namely the case for the models that use linear regression and SVMs. The other end represents the so-called ensemble methods, in which different features are fed to different classifiers and then the resulting decisions are combined to obtain a final decision, as in AdaBoost. In terms of how the feature integration is performed, these two schemes can be categorized as \textit{early integration} and \textit{late integration}, respectively. In that respect, as argued in~\cite{Noble:2004}, the MKL approach provides an \textit{intermediate integration} strategy by delivering a good balance between making too many independence assumptions (as in late integration) versus allowing too many dependencies (as in early integration).

Another key contribution of our work is to employ a large set of  high-level object-specific features as compared to limited number of features (faces, pedestrians, cars) previously utilized in other studies. As will be explained in the subsequent section, these features provide a top-down modulation of object-based attentional selection. In addition to those high-level features, we also include some mid and low-level visual features, which have been shown to be correlated with visual saliency, as in~\cite{Judd:ICCV2009,Borji:CVPR2012}.


\subsection{Objects as high-level features}
\label{ssec:objectbank}
As we mentioned before, recent studies on visual attention have shown that top-down object information are strongly correlated to visual saliency (Fig.~\ref{fig:highlevel}). Either eye fixations of viewers correspond to interesting objects in images~\cite{ElazaryJoV2008}, or top-down object information is solely enough to make accurate saliency predictions~\cite{EinhauserJoV2008}. Even in the literature, it has been argued that objects can be considered as units of attentional selection~\cite{Scholl2001,NuthmannJoV2010}. This been said, the previous saliency models used only a small set of object-specific top-down features. The most commonly used object classes are faces and pedestrians as humans have a tendency to look them in images. Here, our goal is to exploit a more comprehensive set of features which depends on a large group of object classes, and let the learning algorithm extract the correlations among different object classes and visual saliency in an automatic way. 

For our top-down object-specific feature set, we use the ObjectBank representation of images recently proposed by Li et al.~\cite{Li-ObjectBank:NIPS2010}. This representation brings together responses of pre-trained classifiers for a large set of diverse of object classes in a scale invariant way. In the experiments, we used 177 different object detectors in 5 different scales\footnote{We used the code available online at \url{http://vision.stanford.edu/projects/objectbank/}.}, giving us a higher dimensional object-based representation of images.
	
\subsection{Low and mid-level features}
\label{ssec:lowandmid}
In addition to the high-level object-specific features mentioned in the previous section, we have also used the following low and mid-level features:
\begin{itemize}
\item\textbf{Intensity, orientation and color  contrast.} These basic image features are well known for being biologically plausible and being quite effective in determining visual saliency so they are commonly utilized by many saliency methods. Here we directly include the corresponding three channels of these  features as calculated by Itti-Koch saliency model~\cite{Itti:TPAMI1998}.

\item \textbf{Subband Features.} We used the local energy of the steerable pyramid filters~\cite{Simoncelli:ICIP1995}, which are computed in four orientations and three scales, giving us 13 features in total.

\item\textbf{Color Features.} We included values of three color channels (red, green and blue) and the probabilities of each color channel at five different blur level by computing 3D histograms, which in turn provides 11 features.

\item\textbf{Horizon line.} As in~\cite{Judd:ICCV2009,Borji:CVPR2012}, we utilized the response map of a horizon line detector trained over mid-level gist features~\cite{Oliva:IJOCV2001}. The underlying reason behind using this feature is that humans instinctively investigate horizon line for salient objects.

\item\textbf{Bottom-up saliency maps.} We used saliency maps of three different bottom-up saliency models. first, we make use of the saliency features described by Oliva and Torralba~\cite{Oliva:IJOCV2001} and Rosenholtz~\cite{Rosenholtz:1999} which is based on subband pyramids. Second, we use saliency maps obtained with the graph-based saliency model of Harel et al.~\cite{Harel:NIPS2006}\footnote{The source code is downloaded from~\url{http://www.klab.caltech.edu/~harel/share/gbvs.php}}. Lastly, we included saliency map of a recently proposed bottom-up saliency model~\cite{Erdem:JoV2013} referred to as CovSal\footnote{The source code is publicly available at~\url{http://web.cs.hacettepe.edu.tr/~erkut/projects/CovSal/}}, which employs region covariances to encode relationships between some simple visual features such as Intensity, color and orientation.

\item\textbf{Center map.}
Experimental evidence suggests that people have a tendency to look towards image centers. To implicitly incorporate this center bias to our computations, we added a feature map which gives more importance to locations closer to image centers.
\end{itemize}
\section{Experimental Results}
\label{sec:experiments}
\subsection{Eye fixation datasets}
We trained our model on the dataset introduced by Judd et al.~\cite{Judd:ICCV2009}, which we refer to the MIT1003 dataset. This dataset includes a total of 1003 images (779 landscape and 228 portrait images), which were randomly crawled from Flickr and LabelMe. The images in this dataset are mostly of $1024\times 768$ pixels (the longest dimension is 1024 pixels and the other one ranged from 405 to 1024 pixels). The dataset contains eye fixation data from 15 viewers who performed a 3-sec-long free-viewing task on each image. Additionally, we have used the dataset from~\cite{Bruce:NIPS2005}, referred to as the Toronto dataset, which contains 120 natural color images, each depicting an outdoor or an indoor urban scene. The size of the images in this dataset are all the same ($681\times 511$ pixels) and the eye movement data were collected from 20 subjects who free-viewed each image for 4 secs. 

\subsection{Evaluation}
To quantitatively analyze the performance, we employed the commonly used area under the receiver operator characteristics curve (AUC) score. That is, the saliency map is thresholded and treated as a binary classifier on every pixel in a given image such that those pixels with saliency values greater than the applied threshold level is classified as fixated; whereas, the rest are considered as nonfixated~\cite{Tatler:VisionResearch2005}. The fixation data from the subjects are used as ground truth. Then, the consistency between a saliency model and the set of human fixations is given by AUC, obtained by varying the threshold level. An AUC of 1 indicates a perfect prediction, while the chance performance is around an area of 0.5.

\subsection{Experimental setup}
We tested our MKL-based saliency estimation approach with three different MKL approach mentioned in Sec.~\ref{sec:mkl}, namely RBMKL~\cite{Cristianini2000}, NLMKL~\cite{Cortes:NIPS2010} and LMKL~\cite{Gonen:ICML2008}\footnote{We used the publicly available implementations found at \url{http://users.ics.aalto.fi/gonen/jmlr11.php}}. We also implemented two other learning based methods, one based on linear SVM\footnote{We used the Liblinear package available at \url{http://www.csie.ntu.edu.tw/~cjlin/liblinear}} as in~\cite{Judd:ICCV2009} and the other based on AdaBoost\footnote{We used the Gentle AdaBoost library available at \url{http://graphics.cs.msu.ru/en/science/research/ machinelearning/adaboosttoolbox}} as in~\cite{Zhao:JoV2012,Borji:CVPR2012}. For a fair comparison, we kept the setup described below the same for all the approaches evaluated in the experiments. 

One particularly important aspect of learning-based saliency estimation methods is how training 
is conducted. Here, we follow a strategy similar to the one employed by Judd et al.~\cite{Judd:ICCV2009}. We randomly divided the dataset into 903 training and 100 testing images. From each training image, we select 10 positive samples (pixels) from the top 20\% salient regions and 10 negative samples from the bottom 70\% salient regions according to the ground truth eye fixation density maps. Since the sizes of the images in the dataset vary, the aforementioned sampling process is performed after the images are resized to $200\times 200$ pixels. Once we determined the best performing MKL approach for saliency estimation, we have trained another model by using all the images and fixation data in the MIT1003 dataset, and then tested its performance on Toronto dataset.

Moreover, we have trained two different saliency models for each learning approach to explicitly analyze the use of high-level object-specific features for visual saliency estimation. While the first group is based on all the low, mid and high-level features (a total of 917 different features, 32 low and mid-level cues described in Sec.~\ref{ssec:lowandmid} and $177\times 5=885$ object-based high-level features described in Sec.~\ref{ssec:objectbank}), the other group uses only the proposed ObjectBank-based features (just 885 top-down object features). In all cases, the optimal parameters are determined according to the average classification performance based on repeating the experiments 10 times with different randomly chosen partitions.

\subsection{Performance}
\begin{figure}[!t]
\centering
\begin{tabular}{p{0.24\linewidth}@{$\;$}p{0.24\linewidth}@{$\;$}p{0.24\linewidth}@{$\;$}p{0.245\linewidth}}
$\;\quad$ Input images & $\;\;\quad$Eye fixations & $\;\;\quad$ Full features & $\;$Object-level features\\
\end{tabular}
\includegraphics[width=\linewidth]{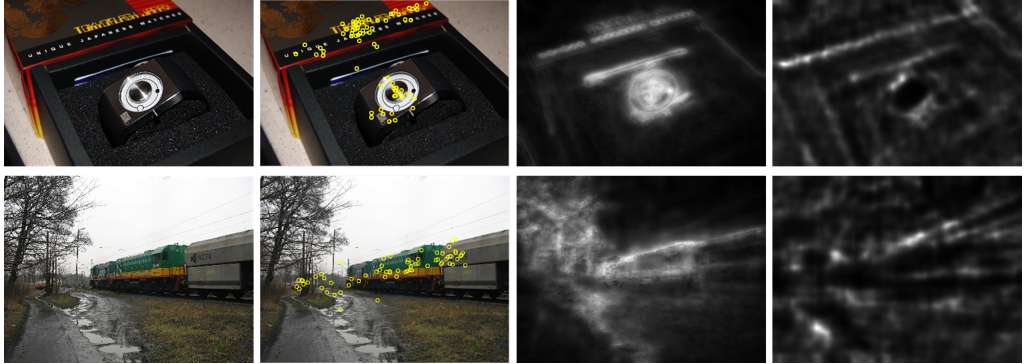} 
\caption{Sample saliency predictions using RBMKL with full vs. object-level features on the MIT1003 dataset. For better illustration, the maps are passed through an exponential function.}
\label{fig:sample_results}
\end{figure}

In the first set of experiments, we compared linear SVM, AdaBoost, NKMKL, RBMKL and LMKL-based saliency learning approaches on MIT1003 dataset using the setup described in the previous subsection. As illustrated in Table~\ref{table_MIT1003}, MKL-based models mostly outperformed all other models by means of AUC score. In particular, the RBMKL-based nonlinear integration is found to be superior to all other feature integration strategies. Another key observation is that the models which use only object-specific features gave rise to worse predictions than the models that also consider low and mid-level features. This suggests that using bottom-up visual information is quite important for saliency predictions, which are inline with the previous studies. However, it is also important to note that with MKL-based models we achieved AUC scores 0.71-0.74 as compared to 0.58 with linear SVM and 0.64 with AdaBoost. This clearly shows that intermediate-level feature integration using MKL is more effective than early and late integration respectively by linear SVM and AdaBoost. In Fig.~\ref{fig:sample_results}, we provide saliency maps obtained with RBMKL for some sample images using two different sets of features.

\begin{table}[!t]
\centering
\caption{AUC Performance comparisons of the learning methods on the MIT1003 dataset (over 10 runs).}
\begin{tabular} {l>{\centering}m{0.14\textwidth}>{\centering}m{0.14\textwidth}>{\centering}m{0.14\textwidth}>{\centering}m{0.14\textwidth}m{0.14\textwidth}}
\hline
& Linear SVM & AdaBoost & NLMKL & RBMKL & $\quad$LMKL \\
\hline
Full features\\
$\quad$\textit{Average} & 0.823 & 0.810 & 0.826 & \textbf{0.835} & $\quad$0.818 \\
$\quad$\textit{Std. dev.} & 0.004 & 0.004 & 0.003 & 0.004 & $\quad$0.005 \\
$\quad$\textit{Best} & 0.830 & 0.816 & 0.833 & \textbf{0.841} & $\quad$0.825 \\
\hline
Object-level features\\
$\quad$\textit{Average} & 0.582 & 0.633 & 0.733 & \textbf{0.739} & $\quad$0.710 \\
$\quad$\textit{Std. dev.} & 0.007 & 0.006 & 0.007 & 0.006 & $\quad$0.021 \\
$\quad$\textit{Best} & 0.592 & 0.645 & 0.750 & \textbf{0.752} & $\quad$0.742\\
\hline
\end{tabular}
\label{table_MIT1003}
\end{table}

\begin{figure}[!t]
\centering
\begin{tabular}{p{0.24\linewidth}@{$\;$}p{0.24\linewidth}@{$\;$}p{0.24\linewidth}@{$\;$}p{0.245\linewidth}}
$\;\quad$ Input images & $\;\;\quad$Eye fixations & $\;\;\quad$ Full features & $\;$Object-level features\\
\end{tabular}
\includegraphics[width=\linewidth]{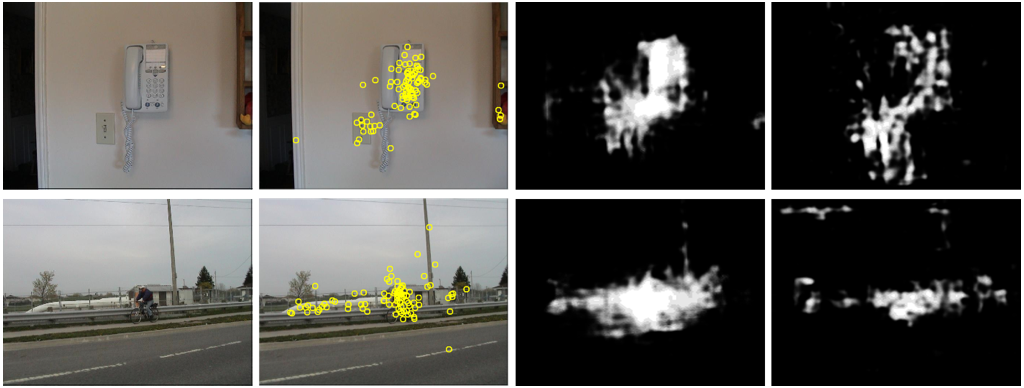} 
\caption{Saliency predictions using RBMKL with full vs. object-level features on the Toronto dataset. For better illustration, the maps are passed through an exponential function.}
\label{fig:sample_results2}
\end{figure}

Additionally, we carried out a second set of experiments on the Toronto dataset, this time with the models learned on the full MIT1003 dataset. Table~\ref{table_TORONTO} shows our quantitative results. As in the first set of experiments, the RBMKL-based models again performed better than the others. Fig.~\ref{fig:sample_results2} demonstrates the results of RBMKL with full and object-level features, and again employing full set of features results in more accurate predictions. In Table~\ref{table_TORONTO}, we also report the AUC scores of the Itti et al.'s bottom-up saliency model~\cite{Itti:TPAMI1998}. Surprisingly, our RBMKL-based model which uses only ObjectBank-based high-level features gave an AUC score quite close to this classical model. 

\section{Discussion}
\label{sec:discussion}
One key difficulty in saliency estimation is to determine the contributions of different visual features to the overall saliency. One of the recent trends in the literature is to estimate these levels of contributions through machine learning schemes by training over ground truth eye fixations from different subjects. In this study, we have also addressed this issue, and suggested to use multiple kernel learning (MKL) to automatically find linear or nonlinear relations between different visual features which are then used in integrating them at an intermediate level. Secondly, we have proposed to enrich the set of visual features with the responses of a very large set of object-specific ObjectBank filters~\cite{Li-ObjectBank:NIPS2010} to model top-down modulations of selective attention, which, up to our knowledge, has never been investigated in this scope before.

\begin{table}[!t]
\centering
\caption{AUC Performance comparisons of the learning methods on the TORONTO dataset.}
\begin{tabular}{m{0.145\textwidth}>{\centering}m{0.14\textwidth}>{\centering}m{0.14\textwidth}>{\centering}m{0.14\textwidth}>{\centering}m{0.14\textwidth}m{0.14\textwidth}m{0.1\textwidth}}\hline
& Linear SVM & AdaBoost & NLMKL & RBMKL & $\quad$LMKL & $\quad\;$Itti \\
\hline
Full features & 0.821 & 0.808 & 0.821 & \textbf{0.831} & $\quad$0.821 & \multirow{2}{*}{$\quad$0.741} \\
Object-level features & 0.591 & 0.627 & 0.730 & \textbf{0.736} & $\quad$0.686 & \\
\hline
\end{tabular}
\label{table_TORONTO}
\end{table}

Our experiments have demonstrated that MKL-based learning schemes give superior performance as compared with the previously proposed linear SVM and AdaBoost-based approaches. As we discussed in Sec.~\ref{sec:approach}, the MKL approach provides an \textit{intermediate integration} strategy~\cite{Noble:2004} by not making too many independence assumptions as in a late integration strategy and not allowing too many dependencies as in an early integration strategy. 

It should be noted that the proposed ObjectBank-based features are highly dependent to detector responses. Since these detectors are not perfect, they may provide some false positives or false negatives, which may, in turn, affect the training and thus the prediction stages in a negative way. Despite this problem, our results have shown that our RBMKL-based model which uses only ObjectBank features gave very competitive results with the traditional bottom-up of model of Itti et al.~\cite{Itti:TPAMI1998}. Clearly, one may argue that using these responses as features does not comply with the standard view of attention where attention is described as the information processing bottleneck. It is important to note that these filters are used to provide a coarse, top-down, somewhat noisy description of the image content, and to discover the level of contributions of the object classes during model learning. On the other hand, from a computational point of view, computing the responses of ObjectBank filters can be also considered as the most burdensome step of the proposed model as it requires semantic processing of the whole image content before estimating the image saliency. 

A possible direction for future work could be to investigate learning task-dependent attention models. The proposed framework is quite general in that it only requires ground truth eye fixations as training data. This data could be collected either under free-viewing conditions or from a task specific experiment.

\section*{Acknowledgements}
We thank the reviewers for valuable comments and suggestions. This research is partly supported by The Scientific and Technological Research Council of Turkey (TUBITAK), Career Development Award 112E146.

\bibliographystyle{splncs03}
\bibliography{egbib}
\end{document}